# From Facts to Folklore: Evaluating Large Language Models on Bengali Cultural Knowledge

**Nafis Chowdhury[1]**   **Moinul Haque[1]**   **Anika Ahmed[1]**
Nazia Tasnim[2]   Md. Istiak Hossain Shihab[3]   Sajjadur Rahman[4]   Farig Sadeque[5]
[1]BRAC University, Bangladesh  [2]Boston University, USA  [3]Oregon State University, USA
[4]Adobe, USA,  [5]BRAC University, Bangladesh

{nafis.chowdhury2,moinul.haque,anika.ahmed}@g.bracu.ac.bd
nimzia@bu.edu, shihabm@oregonstate.edu
sajjadur@megagon.ai,farig.sadeque@bracu.ac.bd

## Abstract

Recent progress in NLP research has demonstrated remarkable capabilities of large language models (LLMs) across a wide range of tasks. While recent multilingual benchmarks have advanced cultural evaluation for LLMs, critical gaps remain in capturing the nuances of low-resource cultures. Our work addresses these limitations through a **Bengali Language Cultural Knowledge (BLanCK)** dataset including folk traditions, culinary arts, and regional dialects. Our investigation of several multilingual language models shows that while these models perform well in non-cultural categories, they struggle significantly with cultural knowledge and performance improves substantially across all models when context is provided, emphasizing context-aware architectures and culturally curated training data.[1]

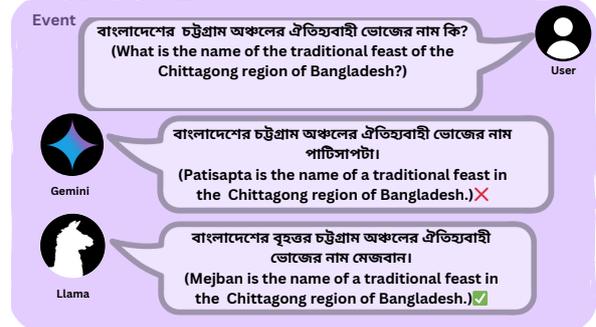

Figure 1: Example of Question Answering from Gemini and Llama in answering Bengali questions (English translations included).

## 1 Introduction and Related Works

Despite the proliferation of Natural Language Processing (NLP) systems as impactful technologies in diverse domains such as education and medicine, systematic inequalities remain in language technology performance across the world's languages (Blasi et al., 2022). While the advancement of NLP technology can often be correlated with popularity, the Bengali language is an oddity - it's the seventh most spoken language globally, with over 250 million speakers (Kawser, 2020), yet remains underrepresented in computational resources and linguistic tools (Tashik et al., 2024). Such resource deficiency poses significant challenges for Bengali Natural Language Processing, as existing toolkits often demonstrate suboptimal performance, even for fundamental tasks such as Parts of Speech (PoS) tagging (Pan and Saha, 2022) and Named Entity Recognition (NER) (Shahgir et al., 2023). The emergence of LLMs has introduced a promising alternative, as these models achieve remarkable performance across a wide range of NLP tasks through zero-shot and few-shot learning (Chowdhery et al., 2023), often outperforming traditional NLP approaches (Ali et al., 2024). However, the effectiveness and reliability of LLMs in Bengali language processing remain unclear (Hasan et al., 2024). Question Answering (QA) is one of the most important tasks in NLP (Farea et al., 2022). Figure 1 illustrates a widely used LLM (Gemini), inaccurately responding to a Bengali cultural question. This gives an intuition about the model's lack of knowledge for the culture.

Within the context of studying LLMs' proficiency in capturing real-world information, recent studies highlight a knowledge gap on retrieval tasks in English, depending on the prominence of entities (Mallen et al., 2022; Maekawa et al., 2024). Moreover, LLMs tend to favor Western or English cultural perspectives, even when operating in other linguistic or cultural contexts (Naous et al., 2024; Wang et al., 2023). However, such considerations remain unexplored for widely used, yet low-resource languages such as Bengali.

---

[1]The dataset and codes are available at https://anonymous.4open.science/r/BLanCK-1E93/

This study investigates the effectiveness of LLMs in Bengali NLP tasks that require cultural and linguistic context. We introduce a dataset to assess LLMs' knowledge and examine whether their characteristics differ from widely used languages. In particular, we explore the following research questions:

- **RQ1.** To what extent do factors such as the popularity of entities impact LLM's knowledge in Bengali?
- **RQ2.** How effectively do LLMs capture Bengali cultural knowledge?
- **RQ3.** How do open and closed LLMs differ in understanding of Bengali cultural knowledge?

## 2 The BLanCK Dataset Curation

### 2.1 Question Answering Task

Past findings show that there is a scarcity of domain-specific QA datasets for the Bengali language (Shahriar et al., 2023). The existing datasets are primarily machine-generated and are prone to biases partly due to ineffective tokenization and inherent limitations that generate surface-level translations (Mahfuz et al., 2024). This work introduces **BLanCK**, a comprehensive evaluation framework combining 5,265 terms from Wikipedia, licensed under CC BY-SA 4.0. We identified 16 diverse semantic categories, grouped into three domains: 44.39% are cultural (event, religion, food, entertainment, historical and people), 42.6% are non-cultural (politics, locations, language, material, organization, health, temporal, economics and sports), and 13.01% are miscellaneous. Each term has two metadata:

- **Context Information:** Contextual descriptions were extracted from Wikipedia for each term, leveraging its structured and verified content to evaluate in-context understanding of LLMs as RAG is under-developed for Bengali (Ipa et al., 2025).
- **Monthly Page Views:** A popularity metric was generated by averaging and normalizing each page's monthly views throughout 2024. This helps examine whether LLMs favor commonly accessed information over culturally significant but less frequently visited content.

To enable systematic evaluation, we used the Qwen-2.5-32B (Qwen, 2024) to generate questions from the context where each term serves as the correct answer. Unlike prior template-based or translated approaches, our method emphasizes contextual understanding of Bengali culture over pattern matching.

### 2.2 Masked Prediction Task

The dataset was also used for masked prediction by replacing the target term with a [MASK] token in its context. If a term appeared multiple times, only the first occurrence was masked. For multi-word terms, each word was masked separately, generating multiple variants per term. This expanded the dataset from 5,265 to 9,114 entries.

## 3 Experiment Setup

In QA, initially, only the question was passed for a brief answer. Secondly, the question was provided with context for a brief answer. The answers were considered accurate only if the term (actual answer) is present as a substring of the generated answer. The models were prompted with the masked contexts for Masked Prediction, and the top five predictions were collected. The models that we used are Gemini 2.0 flash (Google-AI, 2025), Llama3 70B 8192 (AI, 2024), Deepseek V3 0324 (DeepSeek-AI, 2024), GPT4o (OpenAI, 2024), Mistral Small 3.1 (Mistral-AI, 2025b), and Mistral Saba (Mistral-AI, 2025a) for QA. In masked prediction, same models were used except for Mistral Small 3.1, as it failed to respond consistently to a large number of prompts. These models were accessed through API on their default setup by setting temperature at 0, to limit creativity and assess the model's Bengali knowledge. The metrics used for QA and masked prediction are accuracy percentage and MRR, respectively. Table 2 represents performance across both tasks.

## 4 Result Analysis

### 4.1 Impact of Popularity

Figure 2 shows that accuracy increases with term popularity, both with and without context, indicating LLMs have better knowledge of more popular terms. Chunks, grouped into buckets by ascending term popularity, are shown in Figure 2. However, for the most popular chunk with context, accuracy is slightly lower than the second-most popular chunk across all models. Prior work (Mallen et al., 2022) reported lower accuracy with context for the most popular English chunks, but in this case, providing context for Bengali terms improves accuracy, making 'Adaptive Retrieval' unsuitable for Bengali. So, Bengali requires context regardless of term's popularity.

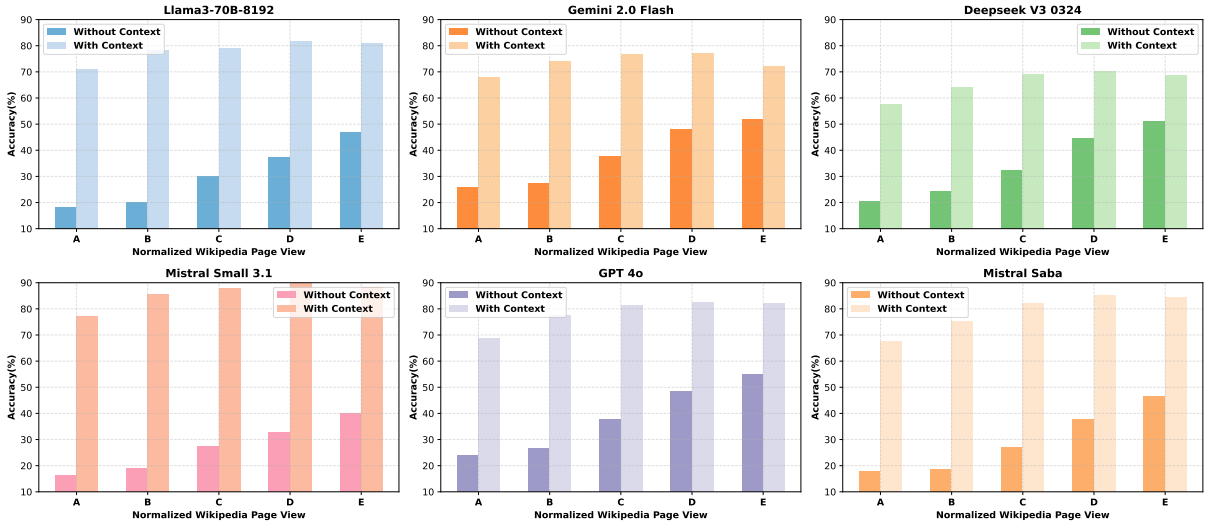

Figure 2: The performance of 6 models with respect to the popularity of terms with and without context. The result for each model was divided into 5 equal chunks (A, B, C, D, and E) by increasing popularity, this buckets terms into popularity groups, and each bar represents the accuracy in that chunk.

| Model | Metric | Misc | Cultural | | | | | | Non-Cultural | | | | | | | | |
|---|---|---|---|---|---|---|---|---|---|---|---|---|---|---|---|---|---|
| | | Misc | Food | Ppl | Rel | Ent | Hist | Event | Pol | Loc | Lang | Mat | Org | Hlth | Temp | Econ | Sprt |
| LLaMA | Acc(%) | 36.99 | 28.46 | 22.49 | 39.03 | **15.23** | 28.17 | 28.81 | 25.26 | 31.95 | 31.82 | 33.33 | 30.74 | 35.83 | 35.24 | 36.84 | **40.70** |
| | MRR | 0.312 | 0.265 | 0.166 | 0.281 | **0.130** | 0.241 | 0.287 | 0.322 | 0.354 | 0.413 | 0.246 | 0.306 | 0.356 | 0.335 | 0.339 | **0.439** |
| Gemini | Acc(%) | 48.32 | 47.10 | 30.99 | 44.27 | **22.69** | 35.21 | 36.44 | 38.34 | 34.97 | 37.91 | 41.79 | 39.81 | **48.33** | 44.76 | 35.09 | 37.65 |
| | MRR | 0.576 | 0.468 | 0.530 | 0.493 | **0.458** | 0.485 | 0.516 | 0.629 | 0.511 | 0.514 | 0.546 | 0.637 | 0.516 | 0.566 | **0.691** | 0.670 |
| DeepSeek | Acc(%) | 43.65 | 38.61 | 26.79 | 42.30 | **16.96** | 35.21 | 33.05 | 31.96 | 33.24 | 30.52 | 37.04 | 36.89 | **45.00** | 39.05 | 28.07 | 43.02 |
| | MRR | 0.487 | 0.493 | 0.424 | 0.449 | **0.275** | 0.433 | 0.506 | 0.562 | 0.512 | **0.582** | 0.479 | 0.509 | 0.499 | 0.511 | 0.562 | 0.571 |
| Mistral Small | Acc(%) | 31.04 | 24.71 | 18.59 | 34.59 | **11.88** | 26.06 | 20.69 | 31.38 | 29.54 | 27.92 | 22.22 | 32.57 | 30.83 | **37.14** | 21.43 | 29.07 |
| | MRR | – | – | – | – | – | – | – | – | – | – | – | – | – | – | – | – |
| GPT | Acc(%) | 46.81 | 51.54 | 29.31 | 41.57 | **24.00** | 27.46 | 34.75 | 35.38 | 37.71 | 32.47 | 45.93 | 42.53 | 47.50 | **53.33** | 36.84 | 41.86 |
| | MRR | 0.452 | 0.460 | 0.363 | 0.374 | **0.303** | 0.342 | 0.407 | 0.519 | 0.457 | 0.518 | 0.415 | 0.475 | 0.490 | 0.469 | **0.582** | 0.560 |
| Mistral Saba | Acc(%) | 35.62 | 35.38 | 23.78 | 35.35 | **16.92** | 23.94 | 27.97 | 28.87 | 28.44 | 24.03 | 32.59 | 31.07 | **41.67** | 36.19 | 31.58 | 33.72 |
| | MRR | 0.083 | 0.023 | 0.057 | 0.016 | 0.029 | 0.053 | 0.024 | 0.054 | 0.073 | 0.073 | **0.014** | 0.053 | 0.071 | 0.048 | 0.047 | **0.093** |

Table 1: The accuracy and MRR per category for every model. The highest and lowest scores of each model are highlighted in bold. From left to right, the categories are Miscellaneous, Food, People, Religion, Entertainment, Historical, Event, Politics, Location, Language, Material, Organization, Health, Temporal, Economics, and Sports. For the Mistral Small model, the masked prediction task was not considered as it failed to respond consistently when processing large portions of the dataset.

### 4.2 Cultural Knowledge Gap in LLM

Cultural context is essential to language comprehension and tasks like NER (Lassen et al., 2023), and language learning benefits from cultural exposure (Genc and Bada, 2005). Model knowledge can be assessed through performance on context-free QA prompts. In both QA and masked prediction, all models performed better on non-cultural terms, as shown in Figure 3, despite dataset having slightly higher number of cultural terms. Domain-specific results show that most models struggled with the cultural 'Entertainment' category, while their best results were in non-cultural domains-except Mistral Saba, which underperformed in a non-cultural category in masked prediction as highlighted in Table 1. This indicates a gap in Bengali cultural knowledge. Moreover, it was crucial to identify par-ticular fields where LLMs underperformed. In QA, models underperformed in most cultural categories except 'Food' and 'Religion', and excelled in all non-cultural categories except 'Economics'. Masked prediction also showed better performance for non-cultural categories, reinforcing this performance gap.

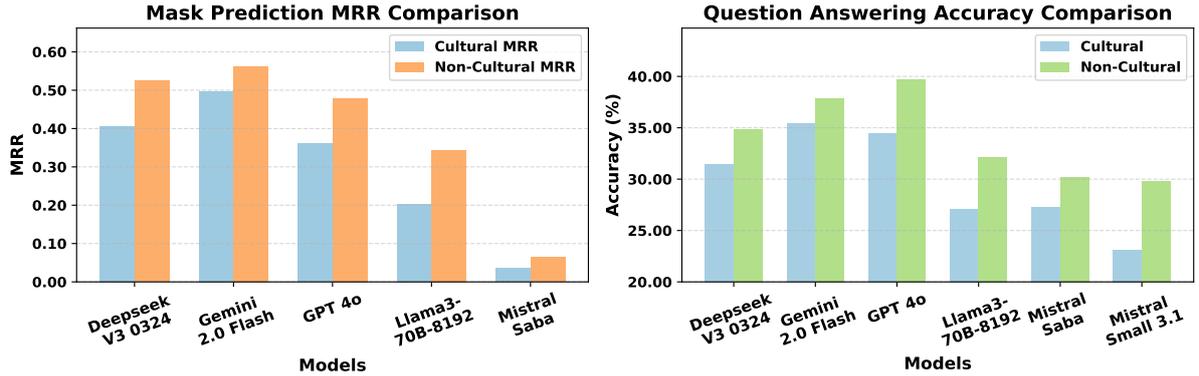

Figure 3: Performance of models for cultural and non-cultural terms for Masked Prediction and QA.

### 4.3 Open and Closed-Source LLM

**Masked Prediction:** As shown in Table 2 Deepseek, an open-source model, outperformed GPT-4o (a closed-source model), but both were surpassed by Gemini, which is the top-performing closed-source model. Except for Mistral Saba, all other closed-source models generally perform better than open-source models for Bengali cultural knowledge.

| Model | Question Answering | | Masked Prediction |
|---|---|---|---|
| | No Context | With Context | MRR |
| Llama3 70B | 30.52% | 78.15% | 0.278 |
| Deepseek V3 | 34.46% | 65.96% | 0.468 |
| Mistral Small 3.1 | 26.97% | 85.61% | — |
| Gemini 2.0 Flash | 38.13% | 73.62% | 0.534 |
| GPT-4o | 38.27% | 78.4% | 0.424 |
| Mistral Saba | 29.56% | 78.85% | 0.053 |

Table 2: Performance of models in Question Answering and Masked Prediction tasks. For the Mistral Small model, the masked prediction task was not considered as it failed to respond consistently when processing large portions of the dataset.

**Question Answering:** Closed models perform better than their open counterparts (Arnardóttir et al., 2025), as seen in the Table 2, where Mistral Saba (closed-sourced) slightly outperforms Mistral Small 3.1 (open-sourced) for prompts without context, despite having equal parameters. Though Mistral Saba is trained on Middle Eastern and South Asian data, its performance is still lower than other models. When prompted without context, closed-sourced LLMs (GPT4o, Gemini) significantly outperformed open-sourced LLMs (Llama, Deepseek), showing better Bengali understanding. The poor performance of open-sourced models points to issues with semantic alignment and deep reasoning (Al Nazi et al., 2025). The results show that open-source LLMs lag behind closed-source LLMs in understanding Bengali culture.

## 5 Conclusion

This work introduces **BLanCK**, a framework for evaluating LLMs on Bengali question answering and masked prediction. Results show that the LLM performance improves with topic popularity and is significantly enhanced by context retrieval, even for well-known terms. Models consistently performed better on non-cultural topics, revealing a gap in understanding Bengali culture and tradition, as LLMs performed better regarding non-cultural categories of information, such as 'sports', 'health', and 'organization'. However, the same models showed poor results when asked for information related to Bengali culture, like 'events', 'entertainment', and 'people'. Closed-source models outperformed open-source ones by a large margin, underscoring the latter's limited Bengali knowledge. Future work should focus on developing culturally rich Bengali datasets that capture diverse aspects such as people, entertainment, and historical events, to enhance contextual understanding and cultural comprehension.

# Limitations

The corpus used in this study is exclusively derived from Wikipedia, a source characterized by formal language and factual content. Alternative data sources, such as social media platforms like Facebook and Twitter (X), were not considered. Notably, X remains relatively unpopular in Bangladesh compared to other regions (The Business Standard, 2024), limiting its utility for constructing a representative Bengali-language dataset. While prior studies (Mallen et al., 2022) have utilized Retrieval-Augmented Generation (RAG) for context extraction, the low-resource nature of Bengali necessitated the use of alternative context-extraction strategies in this work.

Furthermore, the evaluation of language models in this work was limited to question-answering (QA) and masked token prediction. Expanding this scope to encompass additional NLP tasks—such as summarization and next sentence prediction offers a promising direction for future research. Lastly, while this study focused on multilingual models, future investigations could benefit from a deeper exploration of monolingual Bengali models to more precisely evaluate their language-specific capabilities.

# Ethical Consideration

The dataset used in this study comprises exclusively Bengali Wikipedia content, which is publicly available and licensed under Creative Commons (CC BY-SA 4.0). As such, it contains no personal or sensitive information, and no human subjects were involved or harmed. The majority of the content consists of definitions and factual entries. All models evaluated in this research, Gemini 2.0 Flash, LLaMA3 70B, DeepSeek V3, GPT-4o, Mistral Small 3.1, and Mistral Saba were accessed through their official APIs without any fine-tuning. The primary aim of this work is to analyze and highlight the performance gaps of multilingual language models when applied to low-resource languages such as Bengali. All results presented are based solely on model outputs, without any personal interpretations or biases introduced by the authors. AI assistance was only used to enhance the writing.

# References

Meta AI. 2024. Llama 3 model card. https://github.com/meta-llama/llama3/blob/main/MODEL_CARD.md. Accessed: 2024-10-17.

Zabir Al Nazi, Md Rajib Hossain, and Faisal Al Mamun. 2025. Evaluation of open and closed-source llms for low-resource language with zero-shot, few-shot, and chain-of-thought prompting. *Natural Language Processing Journal*, 10:100124.

Nurshat Fateh Ali, Md Mahdi Mohtasim, Shakil Mosharrof, and T Gopi Krishna. 2024. Automated literature review using nlp techniques and llm-based retrieval-augmented generation. *arXiv preprint arXiv:2411.18583*.

Þórunn Arnardóttir, Elías Bjartur Einarsson, Garðar Ingvarsson Juto, Þorvaldur Páll Helgason, and Hafsteinn Einarsson. 2025. Wikiqa-is: Assisted benchmark generation and automated evaluation of icelandic cultural knowledge in llms. In *Proceedings of the Third Workshop on Resources and Representations for Under-Resourced Languages and Domains (RESOURCEFUL-2025)*, pages 64–73.

Damian Blasi, Antonios Anastasopoulos, and Graham Neubig. 2022. Systematic inequalities in language technology performance across the world's languages. In *Proceedings of the 60th Annual Meeting of the Association for Computational Linguistics (Volume 1: Long Papers)*, pages 5486–5505, Dublin, Ireland. Association for Computational Linguistics.

Aakanksha Chowdhery, Sharan Narang, Jacob Devlin, Maarten Bosma, Gaurav Mishra, Adam Roberts, Paul Barham, Hyung Won Chung, Charles Sutton, Sebastian Gehrmann, et al. 2023. Palm: Scaling language modeling with pathways. *Journal of Machine Learning Research*, 24(240):1–113.

Cohere For AI. 2024. c4ai-command-r-plus (revision 432fac1).

DeepSeek-AI. 2024. Deepseek-v3 technical report. *Preprint*, arXiv:2412.19437.

Amer Farea, Zhen Yang, Kien Duong, Nadeesha Perera, and Frank Emmert-Streib. 2022. Evaluation of question answering systems: complexity of judging a natural language. *ACM Computing Surveys*.

Bilal Genc and Erdogan Bada. 2005. Culture in language learning and teaching. *The reading matrix*, 5(1).

Google-AI. 2025. Gemini 2 flash. Accessed April 30, 2025. Available from https://ai.google.dev/gemini-api/docs/models#gemini-2.0-flash.

Md Arid Hasan, Prerona Tarannum, Krishno Dey, Imran Razzak, and Usman Naseem. 2024. Do large language models speak all languages equally? a comparative study in low-resource settings. *arXiv preprint arXiv:2408.02237*.

Atia Shahnaz Ipa, Mohammad Abu Tareq Rony, and Mohammad Shariful Islam. 2025. Empowering low-resource languages: Trase architecture for enhanced retrieval-augmented generation in bangla. In *Proceedings of the 1st Workshop on Language Models for Underserved Communities (LM4UC 2025)*, pages 8–15.

Rumi Kawser. 2020. Bangla ranked at 7th among 100 most spoken languages worldwide. Accessed: 2025-01-28.

Ida Marie S Lassen, Mina Almasi, Kenneth Enevoldsen, and Ross Deans Kristensen-McLachlan. 2023. Detecting intersectionality in ner models: A data-driven approach. In *Proceedings of the 7th joint SIGHUM workshop on computational linguistics for cultural heritage, social sciences, humanities and literature*, pages 116–127.

Seiji Maekawa, Hayate Iso, Sairam Gurajada, and Nikita Bhutani. 2024. Retrieval helps or hurts? a deeper dive into the efficacy of retrieval augmentation to language models. In *Proceedings of the 2024 Conference of the North American Chapter of the Association for Computational Linguistics: Human Language Technologies (Volume 1: Long Papers)*, pages 5506–5521, Mexico City, Mexico. Association for Computational Linguistics.

Tamzeed Mahfuz, Satak Kumar Dey, Ruwad Naswan, Hasnaen Adil, Khondker Salman Sayeed, and Haz Sameen Shahgir. 2024. Too late to train, too early to use? a study on necessity and viability of low-resource bengali llms. *arXiv preprint arXiv:2407.00416*.

Alex Mallen, Akari Asai, Victor Zhong, Rajarshi Das, Daniel Khashabi, and Hannaneh Hajishirzi. 2022. When not to trust language models: Investigating effectiveness of parametric and non-parametric memories. *arXiv preprint arXiv:2212.10511*.

Mistral-AI. 2025a. Mistral saba.

Mistral-AI. 2025b. Mistral small 3.1.

Tarek Naous, Michael J Ryan, Alan Ritter, and Wei Xu. 2024. Having beer after prayer? measuring cultural bias in large language models. In *Proceedings of the 62nd Annual Meeting of the Association for Computational Linguistics (Volume 1: Long Papers)*, pages 16366–16393, Bangkok, Thailand. Association for Computational Linguistics.

OpenAI. 2024. GPT-4o. https://openai.com/index/hello-gpt-4o. Accessed: 2025-04-16.

Subrata Pan and Diganta Saha. 2022. Performance evaluation of part-of-speech tagging for bengali text. *Journal of The Institution of Engineers (India): Series B*, 103(2):577–589.

Qwen. 2024. Qwen2.5: A party of foundation models.

HAZ Shahgir, Ramisa Alam, and Md Zarif Ul Alam. 2023. Banglaconer: Towards robust bangla complex named entity recognition. *arXiv preprint arXiv:2303.09306*.

Md Shihab Shahriar, Ahmad Al Fayad Chowdhury, Md Amimul Ehsan, and Abu Raihan Kamal. 2023. Question answer generation in bengali: Mitigating the scarcity of qa datasets in a low-resource language. In *Proceedings of the 13th International Joint Conference on Natural Language Processing and the 3rd Conference of the Asia-Pacific Chapter of the Association for Computational Linguistics (Volume 1: Long Papers)*, pages 430–441.

Md Iftekhar Islam Tashik, Abdullah Khondoker, Enam Ahmed Taufik, Antara Firoz Parsa, and SM Mahmud. 2024. Advancements and challenges in bangla question answering models: A comprehensive review. *arXiv preprint arXiv:2412.11823*.

The Business Standard. 2024. Title of the Webpage or Article. Accessed: 2025-02-15.

Wenxuan Wang, Wenxiang Jiao, Jingyuan Huang, Ruyi Dai, Jen-tse Huang, Zhaopeng Tu, and Michael R Lyu. 2023. Not all countries celebrate thanksgiving: On the cultural dominance in large language models. *arXiv preprint arXiv:2310.12481*.

# Appendix

## A Dataset Generation

### A.1 Term Extraction

Web scraping was done on Wikipedia to collect the terms of our dataset. As this work focuses on the Bengali Language, the Wikipedia page **"বাংলাদেশ"** ("Bangladesh") was considered to be a source of other important Bengali terms. The Wikipedia page which includes a list information related to **"বাংলাদেশ"** ("Bangladesh") was scraped to extract the titles and links of other Wikipedia pages that are linked to it. This process was done recursively. The titles of the pages collected were considered as the terms of this dataset. Through this method, 5,265 terms were gathered. Additionally, the average monthly page-view per term for the year of 2024 was normalized to be used as a popularity metric.

### A.2 Description of Categories

After the terms were collected, clusters were detected using the Community Detection Algorithm. The clusters were manually inspected, categorized, and merged based on the terms that were present within them. The categories assigned, such as "location" or "sports," were based on identifying common themes or entities that emerged from the terms present in each cluster. The cluster that was diverse and could not be generalized was labeled "miscellaneous". This process resulted in 16 distinct categories: **"location", "miscellaneous", "people", "religion", "entertainment", "organization", "food", "politics", "language", "temporal", "historical", "event", "material", "health", "sports",** and **"economics"**. Among the categories, "people", "religion", "entertainment", "food", "historical", and "events" categories represent **cultural** data since the terms of these categories are associated and exclusive only to Bengali culture and tradition. The rest of the categories, except "miscellaneous," are labeled as **non-cultural** data as the terms present in these categories represent knowledge or entities that are available in almost every culture. A brief description of the categories with examples of terms present in them is given in Table 5.

### A.3 Context Collection and Question Generation

The context was collected for each term from Wikipedia which mostly focused on definition. A set of terms and their corresponding contexts were prompted to both C4AI (Cohere For AI, 2024) and Qwen-2.5-32B (Qwen, 2024) models to generate questions where the target answers would be the terms themselves. There were noticeable differences in the quality of questions generated by each model for specific terms. So, for 50 terms, a lower-scale evaluation was performed to determine the better model for question generation. Three reviewers rated the questions from each model out of 5 based on Relevance (shows how well the question matches the given context or topic), Factual Accuracy (checks if the question is based on correct, error-free information), Question Clarity (measures how unambiguous the question is), Creativity (evaluates originality and depth beyond simple understanding), Coherence (assesses grammar and natural flow of the question), and Task-specific Appropriateness (determines if the question fits the task and context). In all dimensions, Qwen-2.5 outperformed C4AI, for that reason, the rest of the questions were generated by using Qwen-2.5.

| Term | Question | Context | Popularity | Category | Culture Type |
|---|---|---|---|---|---|
| মূর্তি (Statue) | হিন্দু ধর্মে দেবতাদের প্রতিনিধিত্ব করা চিহ্ন বা প্রতিমাকে কি বলা হয়? (What are the symbols or images representing gods in Hinduism called?) | মূর্তি হিন্দু ঐতিহ্যে দেবতা বা মর্ত্যের প্রতিমা বা বিগ্রহের জন্য সাধারণ শব্দ। (Statue is a general term for an idol or statue of a deity or mortal in Hindu tradition.) | 0.00014 | Religion | Cultural |
| চাঁদ (Moon) | পৃথিবীর একমাত্র প্রাকৃতিক উপগ্রহ এবং সৌরজগতে পঞ্চম বৃহত্তম উপগ্রহ কোনটি? (Which is the only natural satellite of Earth and the fifth largest satellite in the solar system?) | চাঁদ পৃথিবীর একমাত্র প্রাকৃতিক উপগ্রহ এবং সৌর জগতের পঞ্চম বৃহত্তম উপগ্রহ (The Moon is Earth's only natural satellite and the fifth largest satellite in the Solar System.) | 0.09147 | Miscellaneous | Miscellaneous |

Table 3: Sample dataset for Question answering with Bengali terms and their English translations in parentheses

| Term | Context | Masked Context | Category | Culture Type |
|---|---|---|---|---|
| মূর্তি (Statue) | মূর্তি হিন্দু ঐতিহ্যে দেবতা বা মর্ত্যের প্রতিমা বা বিগ্রহের জন্য সাধারণ শব্দ। (Statue is a general term for an idol or statue of a deity or mortal in Hindu tradition.) | [MASK] হিন্দু ঐতিহ্যে দেবতা বা মর্ত্যের প্রতিমা বা বিগ্রহের জন্য সাধারণ শব্দ। ([MASK] is a general term for an idol or statue of a deity or mortal in Hindu tradition.) | Religion | Cultural |
| চাঁদ (Moon) | চাঁদ পৃথিবীর একমাত্র প্রাকৃতিক উপগ্রহ এবং সৌর জগতের পঞ্চম বৃহত্তম উপগ্রহ (The Moon is Earth's only natural satellite and the fifth largest satellite in the Solar System) | [MASK] পৃথিবীর একমাত্র প্রাকৃতিক উপগ্রহ এবং সৌর জগতের পঞ্চম বৃহত্তম উপগ্রহ (The [MASK] is Earth's only natural satellite and the fifth largest satellite in the Solar System) | Miscellaneous | Miscellaneous |

Table 4: Sample dataset for masked prediction with Bengali terms and their English translations in parentheses

| Category | Description | Examples |
|---|---|---|
| Location | This category consists of names of places and countries. | ত্রিপুরা, সুন্দরবন, গ্রীস, বোস্টন, ফ্রান্স, ইতালি, তুরস্ক, সিক্কিম, শ্রীলংকা (Tripura, Sundarban, Greece, Boston, France, Italy, Turkey, Sikkim, Sri Lanka) |
| Miscellaneous | The terms that do not fall in any of the other categories. | পেশা, ফিডব্যাক, যন্ত্র, আলিঙ্গন, সত্য, বিজ্ঞান, ভ্রমণ, ব্যবসা, শান্তি, সুফি (Profession, Feedback, Machine, Hug, Truth, Science, Travel, Business, Peace, Sufi) |
| People | It consists of the names of famous people, mostly politicians, writers, singers, and many more, most of whom are from our subcontinent. | সম্রাট জাহাঙ্গীর, জলধর সেন, গগনেন্দ্রনাথ ঠাকুর, শক্তি চট্টোপাধ্যায়, সুকান্ত ভট্টাচার্য (Emperor Jahangir, Jaladhar Sen, Gaganendranath Tagore, Shakti Chattopadhyay, Sukanta Bhattacharya) |
| Religion | This category consists of the names of things related to religion and religious terms. | শিরক, দেবী, হিন্দুধর্ম, সওয়াব, মন্দির, কুরবানী, সালাত, পুরোহিত (Shirk, Goddess, Hinduism, Reward, Temple, Sacrifice, Prayer, Priest) |
| Entertainment | The names of movies, music composer duos, etc. | সাজিদ-ওয়াজিদ, ঘুড্ডি, ভারতীয় চলচ্চিত্র, আলাউদ্দিন আলী, বালাম, জেমস, আলম খান, কাওয়ালি (Sajid-Wajid, Ghuddi (Movie), Indian Cinema, Alauddin Ali, Balam, James, Alam Khan, Qawwali) |
| Organization | It consists of the names of many different types of organizations. | নাসা, গুগল, টুইটার, ইউটিউব, চ্যানেল আই, রেডিও অস্ট্রেলিয়া, দ্য গার্ডিয়ান, ব্র্যাক বিশ্ববিদ্যালয়, (NASA, Google, Twitter, YouTube, Channel I, Radio Australia, The Guardian, Brac University) |
| Food | The names of many different types of food and edible ingredients make up this category. | ল্যাংচা, বাসমতী, পাকোড়া, ফুচকা, জাফরান, আচার (Lyangcha (Sweet), Basmati, Fritters, Fuchka, Saffron, Pickle) |
| Politics | Names of politicians, political parties, and monuments of the Indian subcontinent are included in this category. | মুহাম্মদ আলী জিন্নাহ, গান্ধী, আইয়ুব খান, মুসলিম লীগ, নরেন্দ্র মোদী, শহীদ মিনার (Muhammad Ali Jinnah, Gandhi, Ayub Khan, Muslim League, Narendra Modi, Shaheed Minar) |
| Language | This category includes names of languages. | মারাঠি, ফারসি, উর্দু, তামিল, লাতিন, চীনা (Marathi, Persian, Urdu, Tamil, Latin, Chinese) |
| Temporal | Names of months, seasons, and calendars are present in this category. | হেমন্ত, সেপ্টেম্বর, আষাঢ়, বসন্ত, , জানুয়ারি, নববর্ষ, ঋতু, খ্রিস্টাব্দ (Autumn, September, Ashadha, Spring, January, New Year, Season, AD.) |
| Historical | Terms related to the history of the Indian subcontinent make up this category. | বিজয় সেন, পল্লব রাজবংশ, বঙ্গ রাজ্য, দ্বিতীয় বিগ্রহপাল, সত্রপ, কদম্ব রাজবংশ, কলিঙ্গ, দেবপাল (Vijaya Sena, Pallava dynasty, State of Bengal, Vigrahapala II, Satrap, Kadamba Dynasty, Kalinga, Devapala) |
| Event | This category consists of names of different types of events that happened in our region including military operations, wars, etc. | চীন-ভারত যুদ্ধ, অপারেশন জ্যাকপট, ভারত ভাগ, বাংলাদেশে সামরিক অভ্যুত্থান, কার্গিল যুদ্ধ, (Indo-China war, Operation Jackpot, Partition of India, Millitary Coup in Bangladesh, Kargil War) |
| Material | Names of things we use in our day-to-day lives make up this category. | কাগজ, চেয়ার, স্কার্ট, খেলনা, হেলমেট, নৌকা, চাকা, ভাস্কর্য, পিরামিড, জুতা (Paper, Chair, Skirt, Toy, Helmet, Boat, Wheel, Statue, Pyramid, Shoe) |
| Health | It includes names of diseases, organs, vitamins, etc. | ডেঙ্গু, চোখ, ত্বক, কাশি, মাথাব্যথা, অ্যালার্জি, মৃত্যু, স্তন্যপায়ী, অনিদ্রা, রোগ (Dengue, Eye, Skin, Cough, Headache, Allergy, Death, Mammal, Insomnia, Disease) |
| Sports | Names of players, games, and tournaments make up this category. | ফিফা, ব্যাডমিন্টন, ভলিবল, হকি, দাবা, হ্যান্ডবল (FIFA, Badminton, Volleyball, Hockey, Chess, Handball) |
| Economics | This category includes terms that are related to business, economics or terms that directly influence the economy. | রপ্তানি, আমদানী, টাকা, দুর্নীতি, জিডিপি, মাথাপিছু আয়, মার্কিন ডলার, (Export, Import, Taka, Corruption, GDP, Income per Capita, US Dollar) |

Table 5: Description of 16 Categories with Examples